\documentclass[conference,compsoc]{IEEEtran}
\ifCLASSOPTIONcompsoc
  \usepackage[nocompress]{cite}
\else
  \usepackage{cite}
\fi
\ifCLASSINFOpdf
\else
\fi
\ifCLASSOPTIONcompsoc
 \usepackage[caption=false,font=footnotesize,labelfont=sf,textfont=sf]{subfig}
\else
 \usepackage[caption=false,font=footnotesize]{subfig}
\fi
\usepackage{url}

\usepackage{newunicodechar,graphicx}
\DeclareRobustCommand{\okina}{
  \raisebox{\dimexpr\fontcharht\font`A-\height}{
    \scalebox{0.8}{`}
  }
}
\newunicodechar{ʻ}{\okina}

\usepackage{amsmath}
\usepackage[linesnumbered,ruled]{algorithm2e}

\usepackage{balance}

\begin{document}
\makeatletter
\def\ps@IEEEtitlepagestyle{
  \def\@oddfoot{\mycopyrightnotice}
  \def\@evenfoot{}
}
\def\mycopyrightnotice{
  {\footnotesize
  \begin{minipage}{\textwidth}
  \centering
  \copyright 2021 IEEE. Personal use of this material is permitted. Permission from IEEE must be obtained for all other uses, in any current or future media, including reprinting/republishing this material for advertising or promotional purposes, creating new collective works, for resale or redistribution to servers or lists, or reuse of any copyrighted component of this work in other works.
  \end{minipage}
  }
}
%
\title{Real-Time Event-Based Tracking and Detection for Maritime Environments}



%
\author{\IEEEauthorblockN{Stephanie Aelmore\IEEEauthorrefmark{1},
Richard C. Ordonez, PhD\IEEEauthorrefmark{2},
Shibin Parameswaran, PhD\IEEEauthorrefmark{2} and
Justin Mauger, PhD\IEEEauthorrefmark{2}}
\IEEEauthorblockA{saelmore@hawaii.edu,
richard.c.ordonez.civ@us.navy.mil, \{paramesw, jmauger\}@spawar.navy.mil}
\IEEEauthorblockA{\IEEEauthorrefmark{1}Department of Electrical and Computer Engineering, 
University of Hawai`i at M\=anoa \\
Honolulu, Hawaii 96822}
\IEEEauthorblockA{\IEEEauthorrefmark{2}Naval Information Warfare Center Pacific, 53560 Hull Street,
San Diego, CA 92152}}


\maketitle

\begin{abstract}
Event cameras are ideal for object tracking applications due to their ability to capture fast-moving objects while mitigating latency and data redundancy. Existing event-based clustering and feature tracking approaches for surveillance and object detection work well in the majority of cases, but fall short in a maritime environment. Our application of maritime vessel detection and tracking requires a process that can identify features and output a confidence score representing the likelihood that the feature was produced by a vessel, which may trigger a subsequent alert or activate a classification system. However, the maritime environment presents unique challenges such as the tendency of waves to produce the majority of events, demanding the majority of computational processing and producing false positive detections. By filtering redundant events and analyzing the movement of each event cluster, we can identify and track vessels while ignoring shorter lived and erratic features such as those produced by waves.
\end{abstract}


%
\IEEEpeerreviewmaketitle

\section{Introduction}

\begin{figure*}[!h]
\centering
\subfloat[][Example event frame with annotation labels on all vessels in frame. Stationary vessels are marked 'difficult' since they produce too few events to reasonably be detected. ]{\includegraphics[width=2.2in]{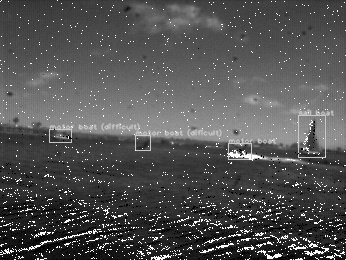}%
\label{fig_frame_177_annot}}
\hfil
\subfloat[][50ms of accumulated events collected and shown on a grey background for illustration. Events are fed into the algorithm asynchronously.]{\includegraphics[width=2.2in]{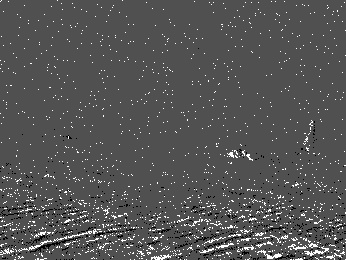}%
\label{fig_frame_177_events}}
\hfil
\subfloat[][For each tracked cluster, both a long-term and short-term velocity are calculated, represented by colored arrows. Confidence scores for each tracked cluster are shown in white. ]{\includegraphics[width=2.2in]{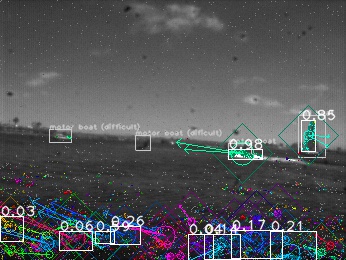}%
\label{fig_frame_177_pmd}}
\caption{An example event frame 8.85 seconds into a DAVIS346red recording after processing. From left to right, three motor boats and a sailboat are visible in frame. The left two motor boats are stationary and therefore are not visible to the stationary event camera and not considered in performance evaluation. The third motor boat is moving to the left and the sailboat is moving to the right. If the difference between the long and short term velocities fora given cluster is very small, the confidence score for that cluster will increase over time}
\label{fig_frame_177}
\end{figure*}

Event cameras (also known as silicon retinas, neuromorphic cameras, or dynamic vision sensors)\cite{patrick2008128x, gallego2019event} are a newer type of digital camera that capture only local changes in intensity rather than capturing a whole frame for every set interval. Since event cameras capture sparse information, they present several advantages for imagery such as increased power efficiency and reduced data bandwidth (see Figure \ref{fig_frame_177_events}). Furthermore, since each pixel operates asynchronously, event-based sensors can capture and process movement without being restricted to a set frame rate, making these sensors ideal for tracking high-speed movement and a good choice for object tracking applications \cite{litzenberger2006embedded}.

When tackling the problem of detecting and tracking vessels in a maritime environment, unique challenges arise for event cameras. In particular, in a rough sea state the visible waves often produce many times more events than the vessels we are interested in tracking. In this paper, we demonstrate an approach for time-constrained maritime object detection. We do this by distinguishing the clusters collected from events produced by waves from the clusters collected from events produced by vessels while minimizing the processing power spent on waves.

Our approach combines noise filtering and clustering in a computationally efficient and parallelizable system which identifies clusters that are likely to be produced by events from a vessel or other object of interest. The movement patterns of these clusters are then analyzed over time to develop a confidence value corresponding to the likelihood that a particular cluster is produced by a vessel or other object of interest.

The following sections provide a breakdown of the state of the art, our methodology, and a demonstration of our methodology on maritime event data collections made from a DAVIS346red camera.

\section{Background}

The authors recognize that while an event camera will capture less redundant information than a conventional frame-based camera, there is still more information than required for identifying and tracking objects. Rodríguez-Gómez, Juan Pablo, et al. recognized this phenomena and randomly sampled a percentage of incoming events across the entire field of view \cite{rodriguez2020asynchronous}. However, by dividing the field of view into partitions, as in our method, one can limit events processed in especially event-dense areas while still processing a high percentage of events from isolated objects.  

Stoffregen and Kleeman presented an event-based approach to optical flow estimation and segmentation that isolates objects and tracks their motion\cite{stoffregen2018simultaneous}. However, this implementation does not run in real-time and collects events into groups based on flow vector rather than spatio-temporal proximity.  In the maritime domain, objects of interest would be grouped together with some of the dynamic background, making them more difficult to isolate.  

Mitrokhin et al. also presented a deep learning approach to motion estimation from event data\cite{mitrokhin2018event}, but the approach required GPU acceleration to run in real-time, significantly increasing the power consumption required. We provide a solution that can run in real-time on a single-core CPU and can be further accelerated with threading or an FPGA implementation due to its simplicity and highly parallel and localized nature.  

Lastly, Benosman et al. presented a computationally efficient event-based optical flow estimation\cite{benosman2012asynchronous}, but Benosman's approach does not isolate or identify different objects in the scene, leaving that as an additional step. For the maritime domain application, distinguishing different objects is the primary goal, so these prior methods do not provide an ideal solution.

\begin{figure*}[!t]
\centering
\includegraphics[width=7in]{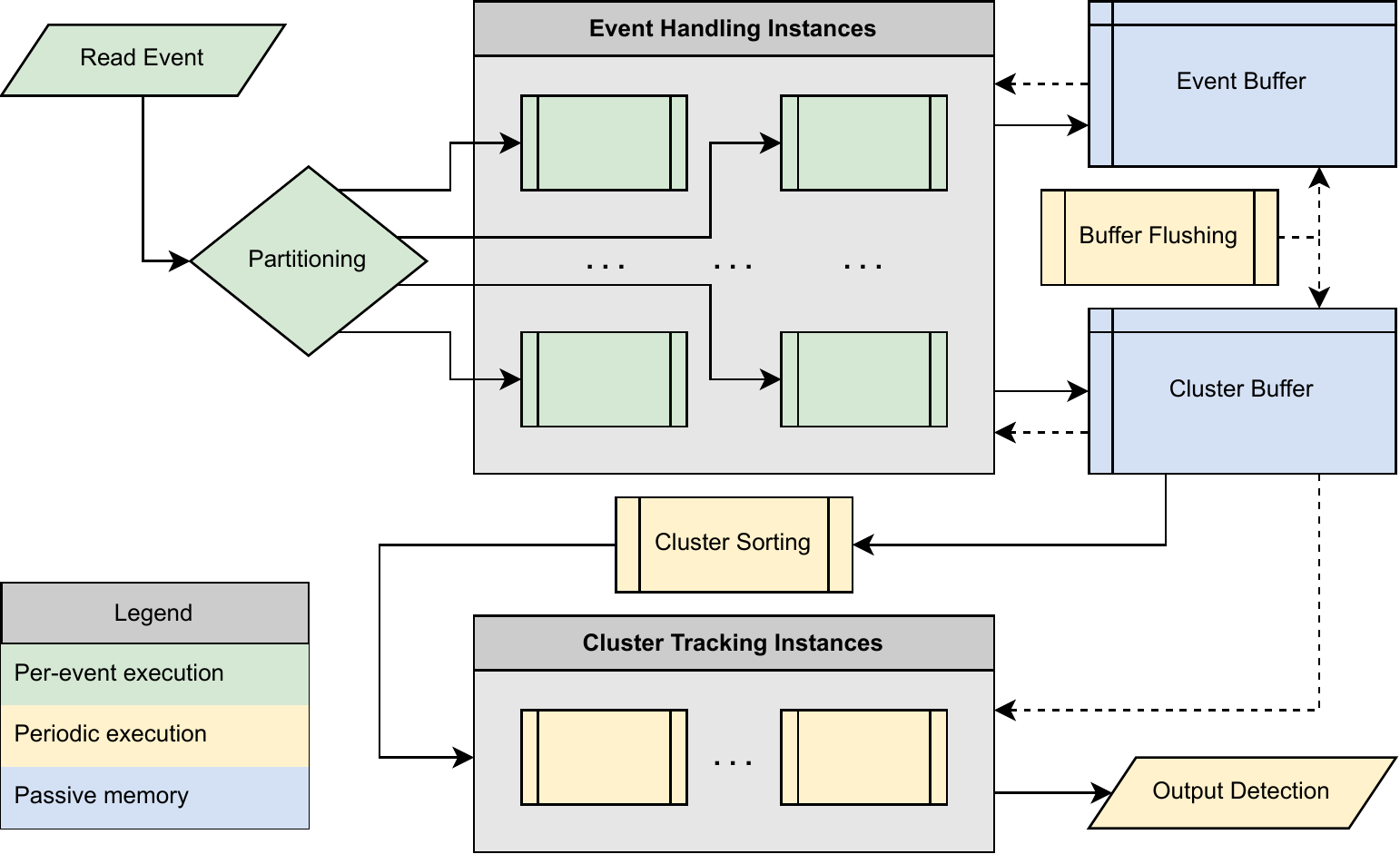}
\caption{An illustration of the overall system architecture. Events are input from the event camera sequentially, then passed to the Partitioning module, where they are sorted by location and assigned to an Event Handling instance. Each Event Handling instance interfaces with the Event Buffer and Cluster Buffer to perform concurrent noise filtering and clustering. The Buffer Flushing module periodically clears expired events from the Event Buffer and updates the Cluster Buffer accordingly. The Cluster Sorting module prioritizes each cluster and assigns the highest priority clusters to be tracked by Cluster Tracking instances. Finally, each Cluster Tracking instance tracks and analyzes the movement characteristics of its assigned cluster and outputs a Detection with a corresponding location, bounding box, and confidence value. The Partitioning and Event Handling modules operate on a per-event basis as each input event is read. The Event Buffer and Cluster Buffer are memory modules that can be written to and read from by other modules. The Buffer Flushing, Cluster Sorting, and Cluster Tracking modules operate periodically.}
\label{fig_flowchart}
\end{figure*}

\section{Architecture}

The overall system is organized as illustrated in Figure \ref{fig_flowchart}. Each event read from the event camera has a corresponding position $(x, y)$, timestamp $t$, and polarity $p$. For our purposes we will not take polarity into account, but $t$ and $(x, y)$ will be passed to the Partitioning module. The system is designed to operate asynchronously and concurrently to the maximum extent possible, with a fixed amount of memory and processing demand.

\begin{figure}[!h]
\centering
\includegraphics[width=3.5in]{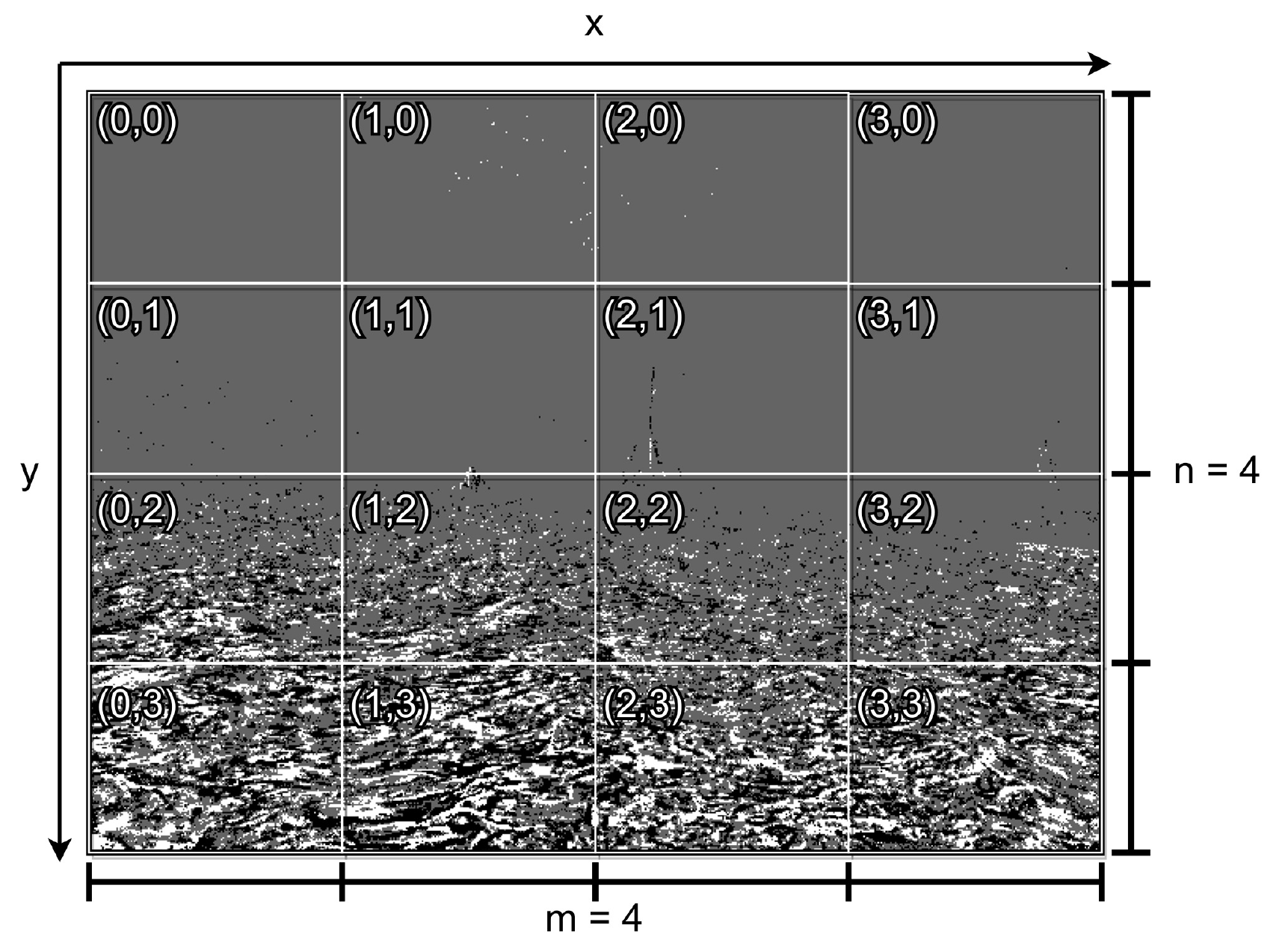}
\caption{An event frame with $m=4$ and $n=4$, resulting in sixteen total partitions. $(x, y)$ coordinates start in the upper left of the image and determine which partition each event is assigned to.}
\label{fig_partition}
\end{figure}
\begin{figure*}[!h]
\centering
\subfloat[][The image field, showing the cluster assigned to the top event in each location as a different color. The light green cluster was collected from events produced by a vessel moving from right to left. The red square indicates the area of the buffer displayed in the plot on the right.]{\includegraphics[width=3in]{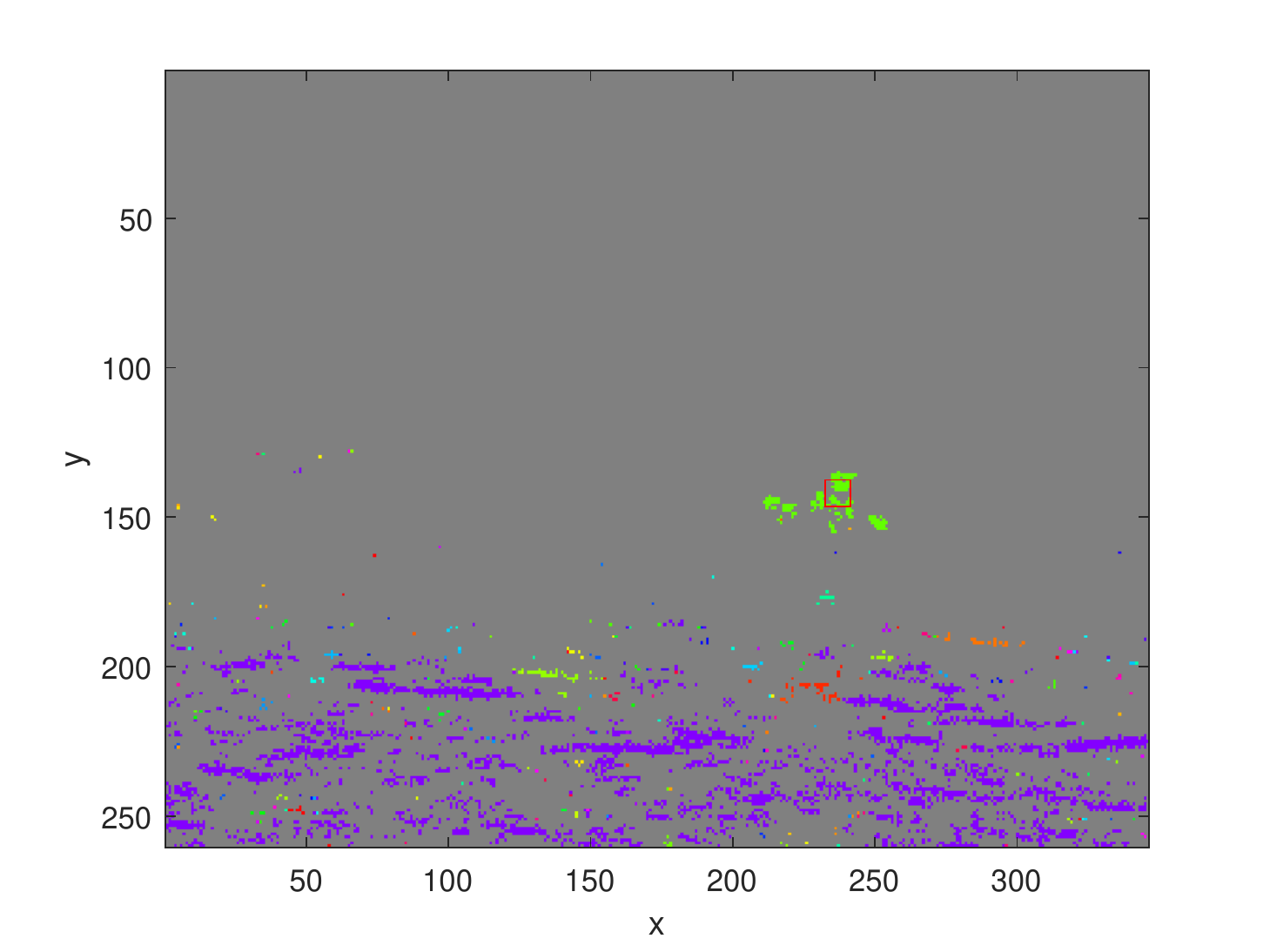}%
\label{fig_frame299_frame}}
\hfil
\subfloat[][A view of the spatio-temporal neighborhood in the Event Buffer surrounding a new event at $(237, 141)$. The red cube represents the spatio-temporal neighborhood of $(x, y)\pm1$ pixels and within $t \pm \tau_f$ or $t \pm \tau_c$ $\mu$s. Only the buffered events inside the red cube are considered for filtering or clustering.]{\includegraphics[width=3in]{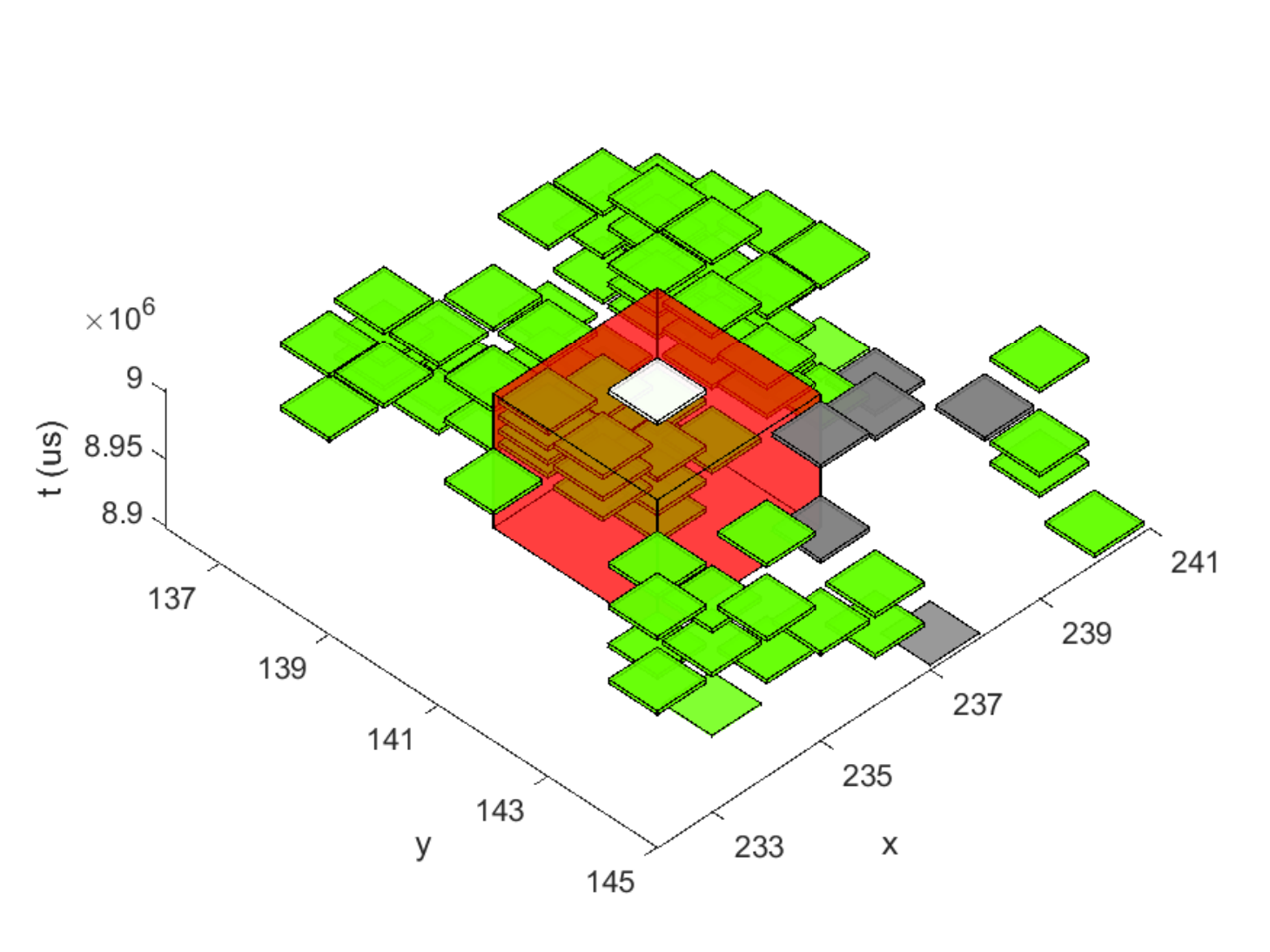}%
\label{fig_frame299_buffer}}
\caption{An illustration of Event Buffer contents at a single point in time.}
\label{fig_frame299}
\end{figure*}

\subsection{Partitioning}
By  dividing  the  field  of  view  into partitions our methods can limit events processed in especially event-dense areas while still processing a high percentage of events from isolated objects. To do this, the image field is divided into $m\times n$ partitions, where $n$ is the number of vertical divisions and $m$ is the number of horizontal divisions. An example partitioning of an event frame can be found in Figure \ref{fig_partition}. Each event is passed to the appropriate Event Handling instance corresponding to the partition in which it falls, according to the value of $(x, y)$.

\subsection{Event Handling}

Each Event Handling instance has a set time per event $\tau_e$ allocated for processing, such that if another event arrives while an event is currently being processed, the incoming event will be ignored and discarded. This allows the rate of events processed to be strictly limited to reduce processing demand.

Each event is first passed through a temporal filter. The top of the Event Buffer at $(x, y)$ is read to obtain the timestamp of the most recent buffered event at that location. If the timestamp is within $\tau_t$ of $t$, then the incoming event will be discarded. Otherwise, the event is passed to the noise filter.

The noise filter counts the buffered events in the spatio-temporal neighborhood as illustrated in Figure \ref{fig_frame299}. The spatio-temporal neighborhood consists of all buffered events from pixels immediately adjacent to $(x,y)$ and occurring within 
a certain time threshold of $t$. 
Events that fall within the \emph{filter threshold} $\tau_f$ of $t$ are counted, producing $n_e$, the number of adjacent events. The set of adjacent clusters $A$ is assembled from clusters assigned to neighboring events within the \emph{clustering threshold} $\tau_c$ of $t$. If $n_e>4$, then the event passes the filter and will be considered for clustering. This last condition ensures that the events which are considered for clustering have temporal persistence, thereby acting as a temporal filter. 

To assign the event to an appropriate cluster, first the adjacent clusters $A$ are sorted by age. The event will be assigned to the oldest adjacent cluster whose centroid is within $d$ pixels of $(x, y)$ by Manhattan distance. If there are no adjacent clusters with centroids within range, then a new cluster will be created consisting of only $\{e\}$.

\subsection{Buffer Flushing}

Periodically, events older than both $\tau_f$ and $\tau_c$ are flushed from the Event Buffer, and the affected clusters in the Cluster Buffer are updated. This maintains a current event count and accurate centroid position for each cluster.

\subsection{Cluster Sorting}

Periodically, all clusters in the Cluster Buffer are sorted by the number of buffered events belonging to each cluster. Clusters with a higher number of events are ranked higher, and clusters with zero events are considered inactive and ineligible for tracking. Sorting is performed using the C++ standard library, with a time complexity of $O(N \log(N))$, where $N$ is the maximum number of clusters. \cite{ISO:2012:III}.

\subsection{Cluster Tracking}

Each Cluster Tracking instance is assigned a cluster of decreasing priority as determined by the Cluster Sorting module. Once a cluster is assigned to a Cluster Tracking instance, it will be tracked until its size has reached zero. While each cluster is being tracked, its centroid position will be recorded periodically. Samples are retained for $T_v$ microseconds. To calculate $\mathbf{v}$, or the long-term velocity, the oldest centroid position $\mathbf{c_v}$ is compared with the current centroid position $\mathbf{c}$ and divided by $T_v$ as shown in Equation \ref{eq_v}. Similarly, the short-term velocity $\mathbf{u}$ is calculated using $\mathbf{c_u}$, a recorded centroid position $T_u$ microseconds in the past, as shown in Equation \ref{eq_u}.

\begin{equation}
    \mathbf{v} = \frac{\mathbf{c}-\mathbf{c_v}}{T_v}
\label{eq_v}
\end{equation}

\begin{equation}
    \mathbf{u} = \frac{\mathbf{c}-\mathbf{c_u}}{T_u}
\label{eq_u}
\end{equation}

\subsection{Detection}

To characterize the motion of each tracked cluster, we periodically compare the long-term and short-term velocities $\mathbf{v}$ and $\mathbf{u}$ using two distinct ratios which are used to add to or subtract from a persistent \emph{stability score} $s$. On every timestep, $s$ is increased by the \textit{difference ratio} $r$, calculated in Equation \ref{eq_diffrat}. Equation \ref{eq_diffrat} places the difference between the two in the denominator so that as the difference between $\mathbf{v}$ and $\mathbf{u}$ decreases, the value of $r$ increases. $\epsilon$ is set to a small value to prevent division by zero in the event that $\mathbf{v}=\mathbf{u}$. This causes the stability $s$ to increase sharply for clusters that hold a steady velocity for even a short amount of time, while still allowing it to accumulate for objects that hold a nearly steady velocity over a longer period.

\begin{equation}
    r = \frac{|\mathbf{v}|}{|\mathbf{v}-\mathbf{u}| +\epsilon}
\label{eq_diffrat}
\end{equation}

To mitigate the constant increase of stability over time, we also subtract a constant multiple of the \textit{angle ratio} $r_a$, which uses the dot product between $\mathbf{v}$ and $\mathbf{u}$ to produce a  dis-similarity measure, as shown in Equation \ref{eq_anglerat}. Unlike $r$ which can grow quite large, $r_a$ is normalized so that it ranges from $0$ (when $\mathbf u=\mathbf v$) to $1$ (when $\mathbf u=-\mathbf v$). 

\begin{equation}
    r_a =\tfrac{1}{2} \left(1- \frac
    {
        \mathbf{v}\cdot \mathbf{u}
    } {
        \sqrt{|\mathbf{v}|^2 \cdot |\mathbf{u}|^2}
    }
    \right) 
\label{eq_anglerat}
\end{equation}

\begin{figure}[h]
\centering
\includegraphics[width=3.5in]{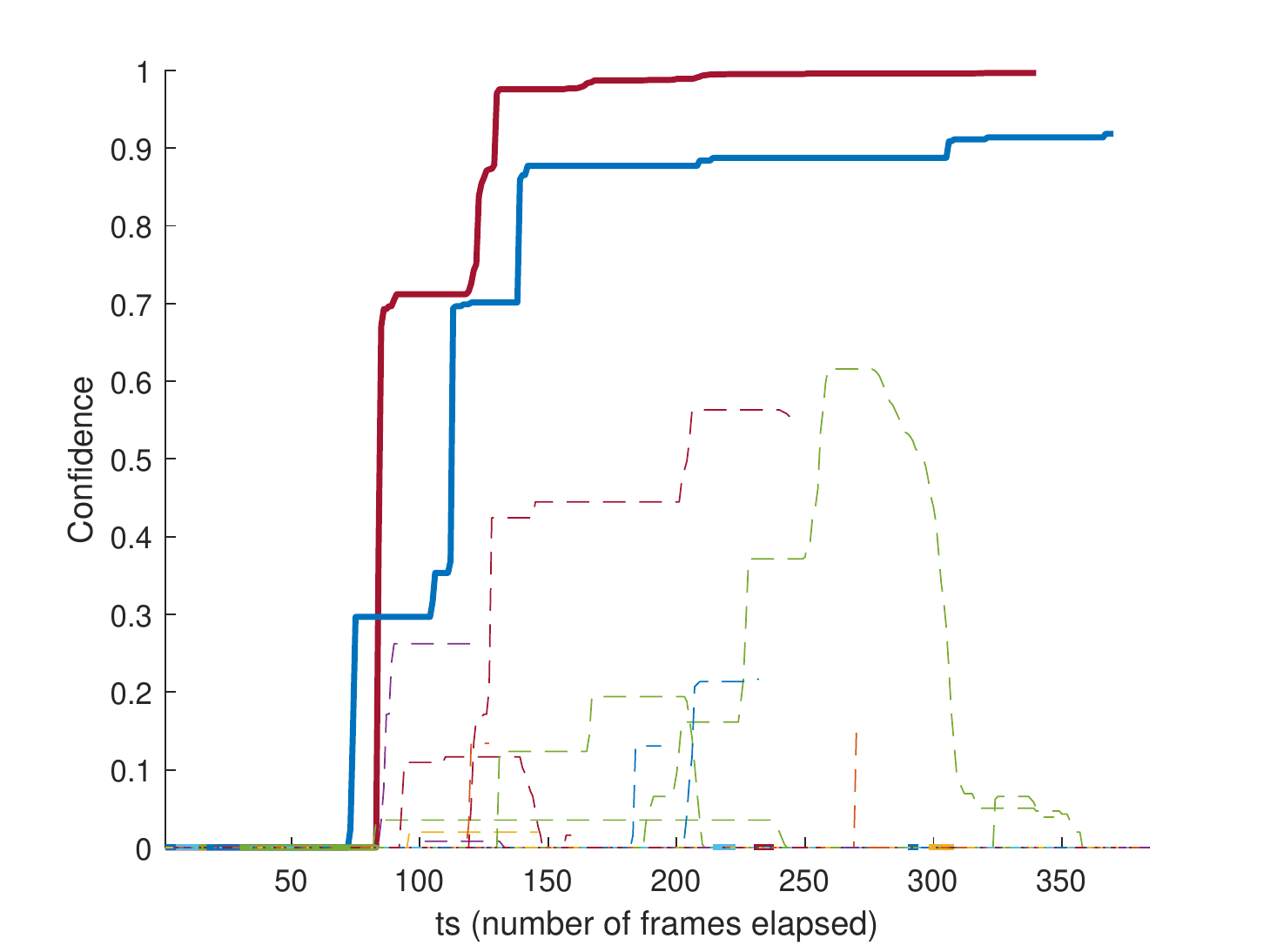}
\caption{Cluster confidence plotted over time for a select video capture from collection 1. The two vessel clusters are shown in bold and the non-vessel ones are shown as dashed lines.}
\label{fig_confovertime}
\end{figure}

At every timestep, a multiple $c_a\cdot r_a$ is subtracted from $s$, where $c_a$ is the \textit{scaling factor} that represents the decrease in $s$ when $\mathbf u$ and $\mathbf v$ are anti-parallel.
The combined change in $s$ each timestep is shown in Equation \ref{eq_stability}, where $i$ is the number of timesteps over which a given cluster has been tracked.

\begin{equation}
    s_{i} = 
    \begin{cases} 
      s_{i-1} + r - c_a r_a & i > 0 \\
      0 & i=0 \\
   \end{cases}
\label{eq_stability}
\end{equation}
Finally, for each detection we produce a confidence score $w$ from $s$ using an exponential function as shown in Equation \ref{eq_exp}. The confidence $w$ ranges from $0$ to $1$ and is therefore suitable for use with standard computer vision metrics such as a receiver-operator curve. Over time, the confidence value tends to go to 1 for clusters exhibiting consistent stable motion, and tends to go negative (effectively 0) for clusters exhibiting erratic motion, as illustrated in Figure \ref{fig_confovertime}.

\begin{equation}
    w = 1-e^{-s}
\label{eq_exp}
\end{equation}

\begin{figure*}[h]
\centering
\subfloat[][Collection 1 shows a peak at 460 scaling factor, followed by a sharp drop. This can be attributed to waves presenting similar motion characteristics to the vessels, at least over short time intervals. Marginal improvement can be gained with an increased scaling factor up to a point, beyond which the stability score of vessels as well as waves starts to decrease, negatively affecting detection performance.]{\includegraphics[width=2.2in]{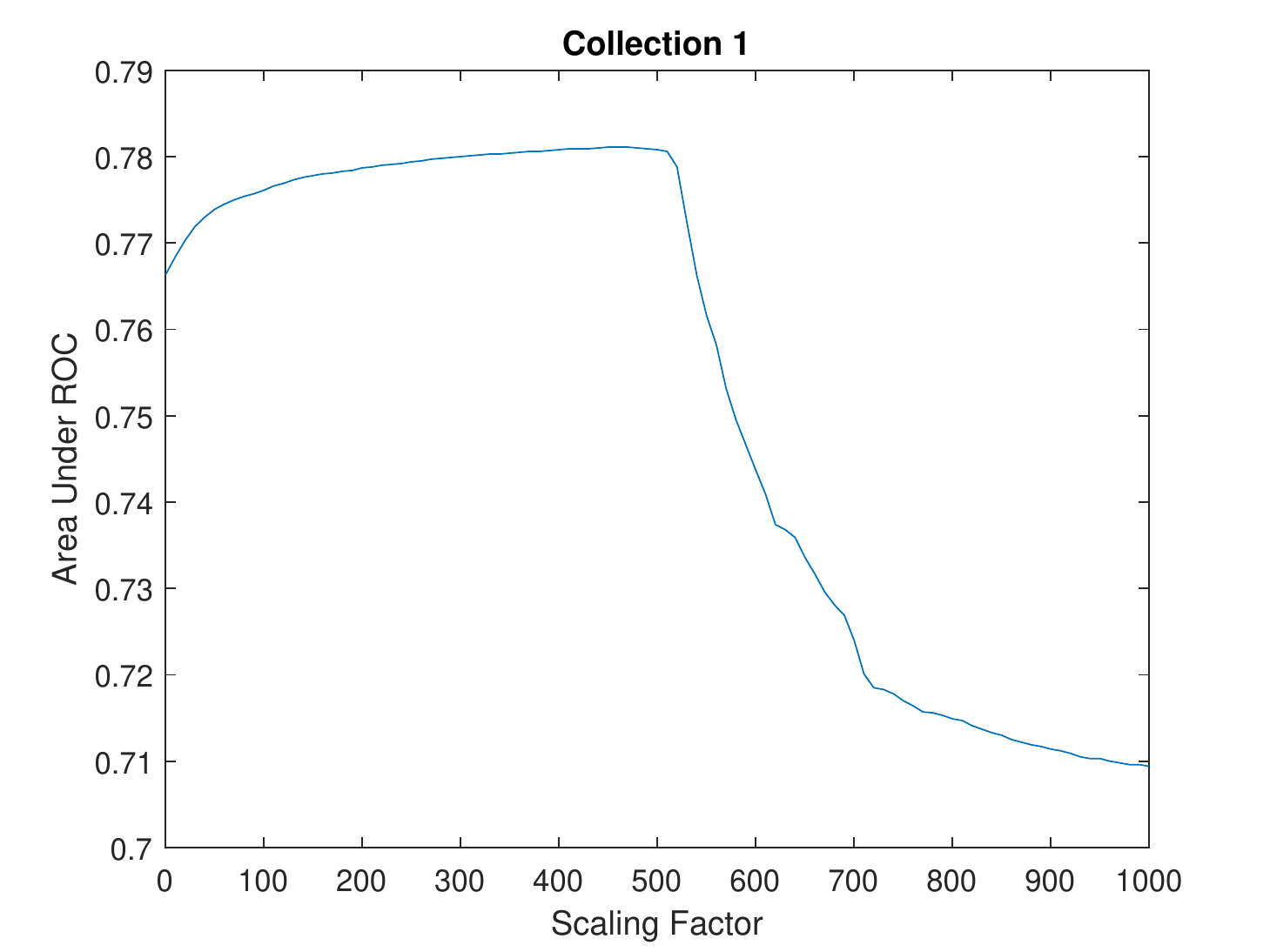}%
\label{fig_scaling_1}}
\hfil
\subfloat[][Collection 2 shows a peak at 760 scaling factor, nearly reaching a plateau before slowly beginning to drop off. In collection 2 there is a high degree of variability between different clusters, with many persistent background clusters produced by the boat wakes. These persistent background clusters move erratically, but need a high enough scaling factor to bring down their stability score after short intervals of persistent motion so that they may be distinguished from targets of interest]{\includegraphics[width=2.2in]{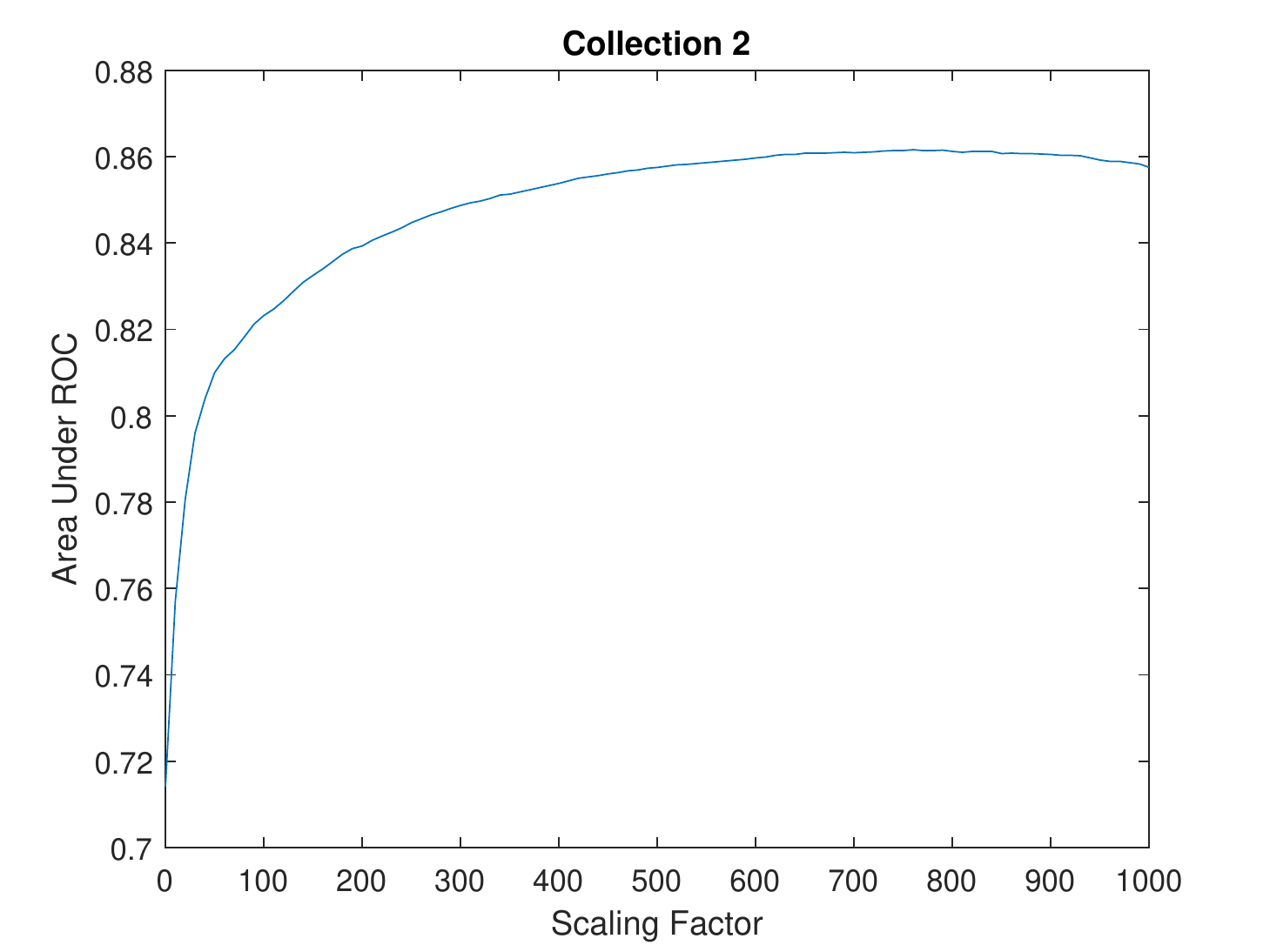}%
\label{fig_scaling_2}}
\hfil
\subfloat[][Collection 3 shows a peak at 0 scaling factor. In collection 3 there is very little dynamic background at all, due to the calm sea and steady camera. For this reason, introducing a non-zero scaling factor decreases the resulting performance albeit by a small amount.
]{\includegraphics[width=2.2in]{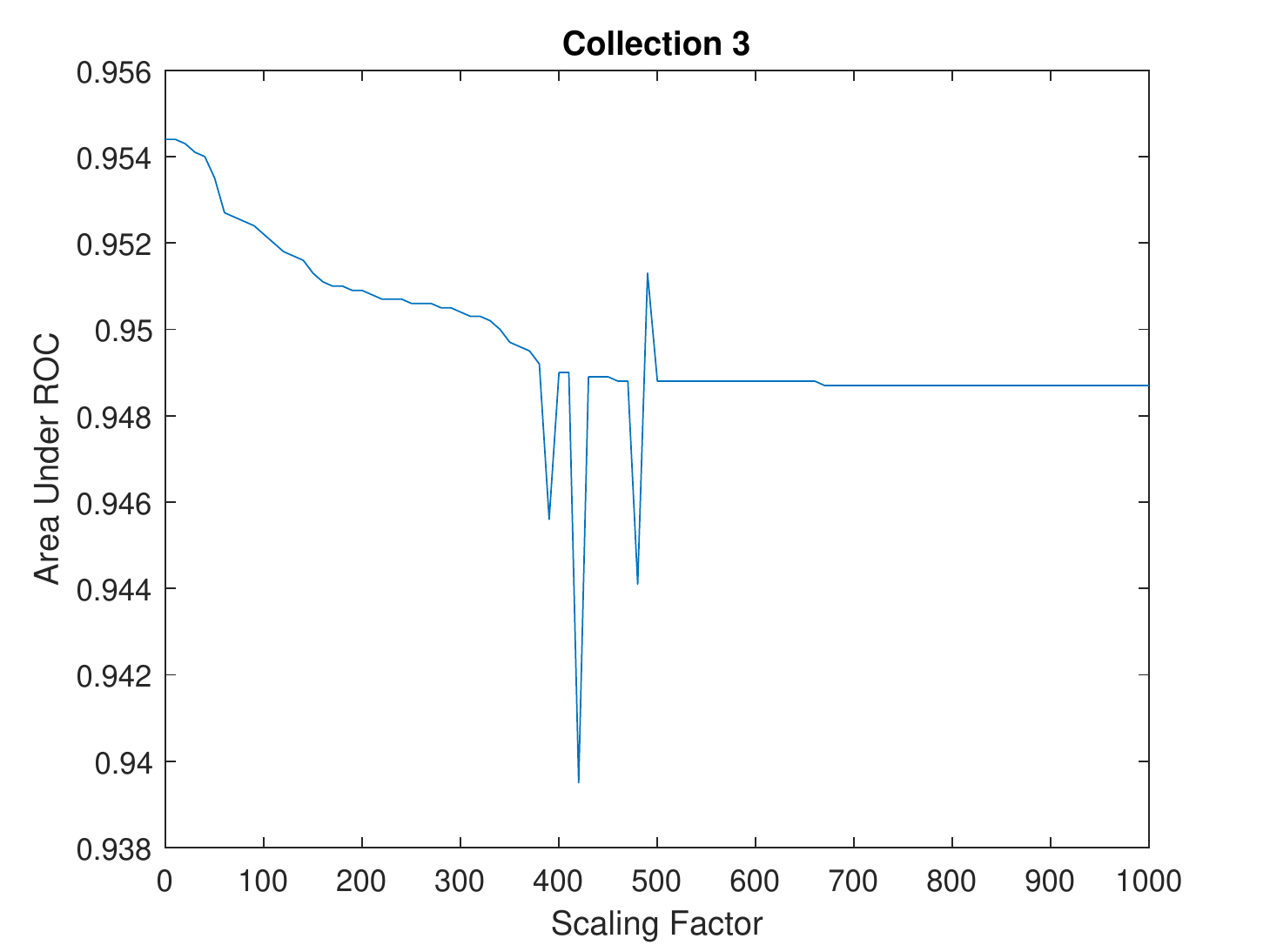}%
\label{fig_scaling_3}}
\caption{Scaling factor ($c_a$) determination results.}
\label{fig_scaling}
\end{figure*}

\begin{figure*}[h]
\centering
\subfloat[][Optimal results obtained on collection 1, with area under the curve of 0.781. True positive rate does not reach 1.0 due to frames for which there are vessels but no clusters within the ground truth bounding box, so they are not considered positive detections at any confidence threshold.]{\includegraphics[width=2.2in]{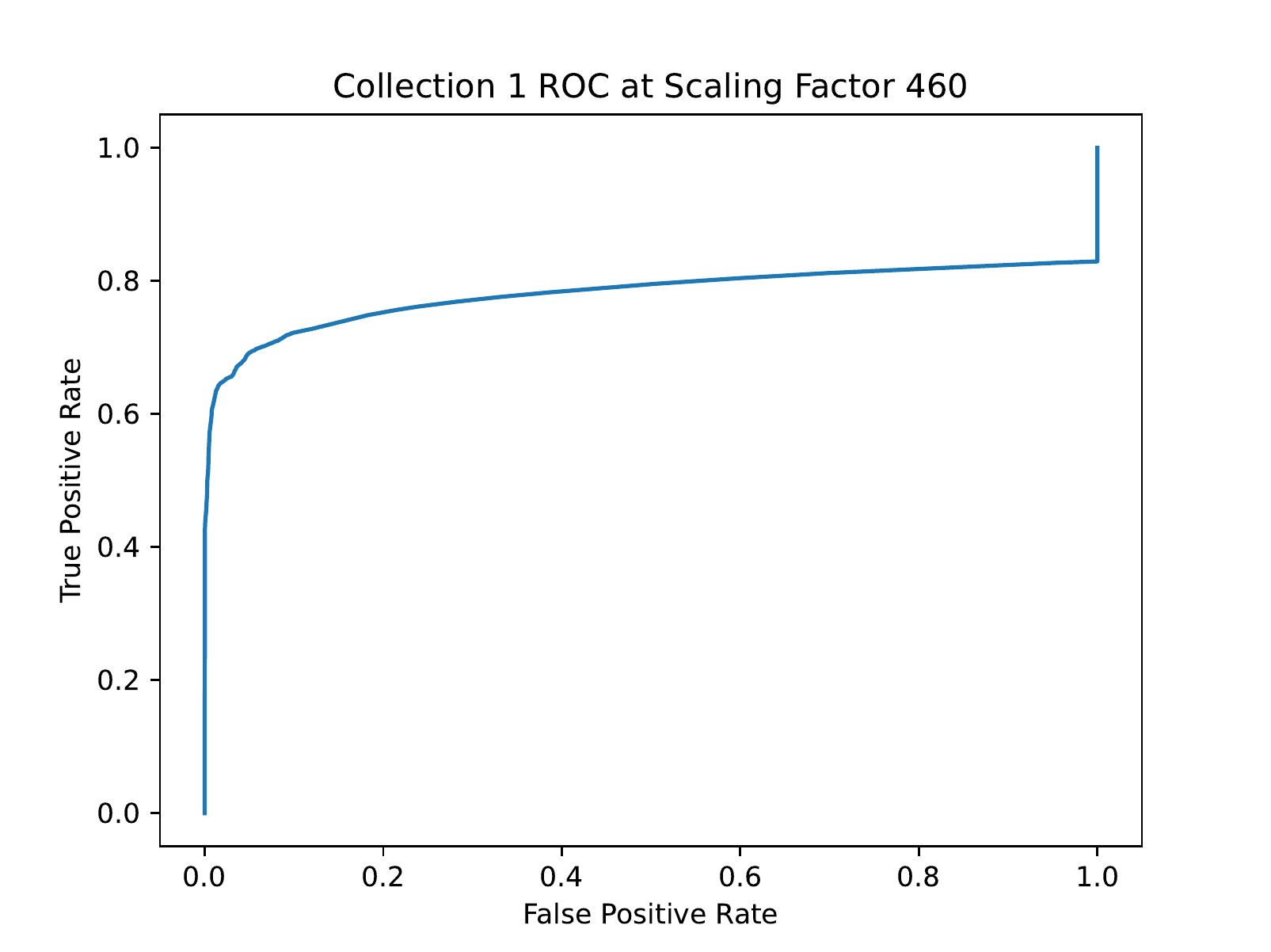}%
\label{fig_roc_1}}
\hfil
\subfloat[][Optimal results obtained on collection 2, with area under the curve of 0.862. Poor performance on this data set is due to the movement of the camera and the wake of the boat producing a high number of false positives.]{\includegraphics[width=2.2in]{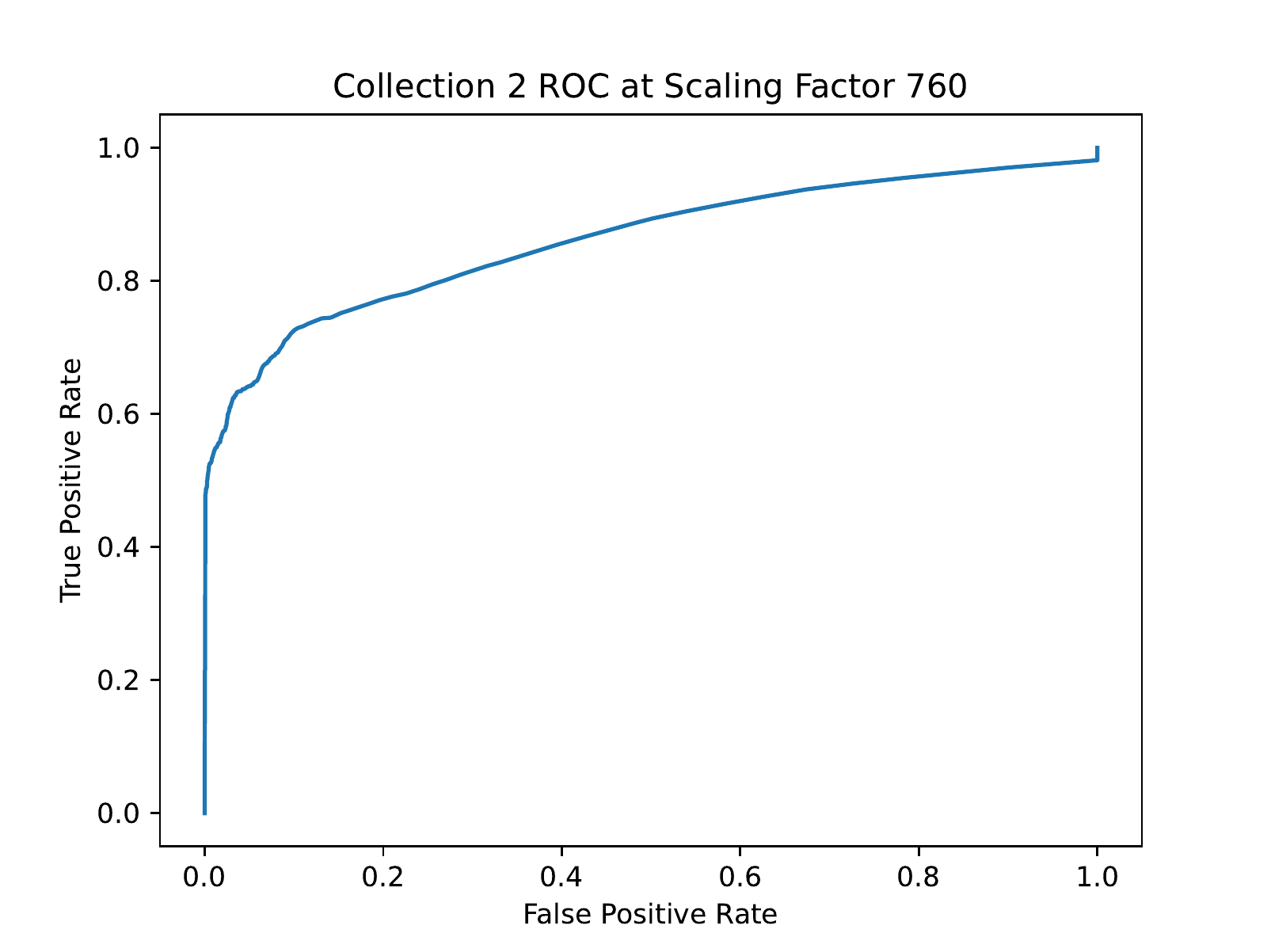}%
\label{fig_roc_2}}
\hfil
\subfloat[][Optimal results obtained on collection 2, with area under the curve of 0.954. A stable camera and limited background motion make collection 3 a near-ideal case for our approach, yielding the highest peak results.]{\includegraphics[width=2.2in]{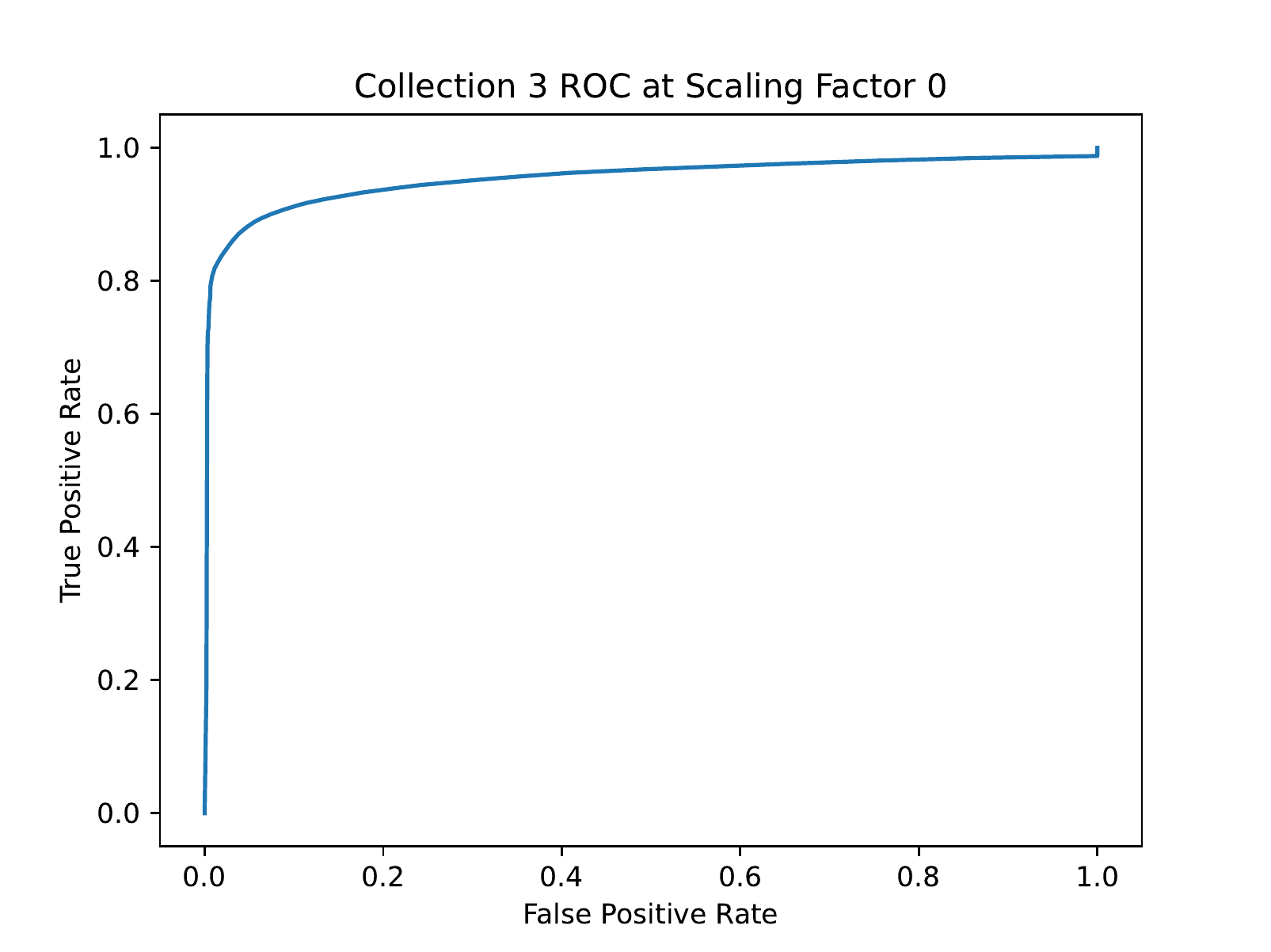}%
\label{fig_roc_3}}
\caption{Receiver-operator curve for the optimal scaling factor for each collection.}
\label{fig_roc}
\end{figure*}

\section{Experimentation}

We illustrate our methodology on maritime data collected using a DAVIS346red camera. Ground truth annotations were applied manually to each frame using the Computer Vision Annotation Tool (CVAT). Annotations with very few corresponding events were marked `difficult' and ignored by subsequent steps. 
Data was collected on three separate occasions of boats moving through a harbor under different conditions. Collection 1 consists of 4 captures totaling 84 seconds with a stationary camera and high sea state. Collection 2 consists of 4 captures totaling 53 seconds with a camera placed on the deck of a moving boat at a high sea state. Collection 3 consists of 7 captures totaling 235 seconds with a stationary camera at long range and a low sea state.

We selected $m=4$ and $n=4$ for 16 total partitions based on the typical size on vessels in the field of view, so that the majority of events belonging to objects of interest would be processed, while some of the events in the background areas would be ignored to save processing time. $\tau_e$ was set to 50 microseconds, which proved to be sufficient time to process the majority of events given the number of partitions and input event rate. $\tau_t$ was set to 100 milliseconds, which proved sufficient to prevent hot pixels from maintaining their own clusters without any correlated events. $\tau_f$ and $\tau_c$ were both set to 200 milliseconds for the sake of simplicity, which seemed to provide a decent balance between filtering out noise and collecting all events from objects of interest into consistent clusters. Maximum cluster size $d$ was chosen to be 30 pixels, since that corresponded to the typical size of vessels in our data. Long term velocity period $T_v$ was 3 seconds, and short term velocity period $T_u$ was 2 seconds. These time intervals provided a good balance between responsiveness and stability.  Optimizing these parameters to fit other hardware and data constraints is left for future work. In this paper, we limit such optimization to a single parameter, namely the scaling factor $c_a$, for which we evaluate the performance of our approach on each collection over a range of values, as shown in Figure \ref{fig_scaling}. 

For each ground truth target on every frame, the detection with the highest confidence that intersects its bounding box was considered to match and added as a true positive. Any additional detection intersecting a ground truth bounding box was ignored. Detections not intersecting any ground truth target were added as false positives. If a ground truth went undetected for any frame it would be added as a false negative.

For each data collection, all accumulated true positives, false positives, and false negatives were considered over a range of confidence thresholds to construct a receiver-operator curve, with the area under the curve used to quantify the effectiveness of our approach. Optimal curves for each data collection can be found in Figure \ref{fig_roc}.

\section{Conclusion}
Our findings indicate that the multi-stage filtering and clustering strategy is successful at ignoring the majority of the spurious events caused by wave movements. The cluster analysis and movement-based confidence scores further refine the detection results by removing shorter-lived and erratic event clusters such as those produced by waves. In summary, the presented algorithm is effective at identifying moving man-made objects in the maritime domain while keeping the computational load to a minimum, making it well-suited for edge-ready applications.

\balance

Future work will focus on developing an FPGA implementation to speed up event data processing. The parameters and methods in this work are optimized to require minimal processing resources, given our specific dataset and hardware constraints. This may not be optimal for other hardware configurations and future data collections. Therefore, automating optimal parameter selection and increased parallelization to handle data collected using higher resolution event imagers are other directions of interest for future research. 



\section*{Acknowledgments}
The authors would like to thank the Veterans To Energy Careers (VTEC) and the Naval Innovative Science and Engineering (NISE) programs for their support.



\balance
\bibliography{IEEEabrv, main.bib}
\bibliographystyle{IEEEtran}
\end{document}